\documentclass[lettersize,10 pt,journal, twoside]{IEEEtran}

\usepackage{graphics} 
\usepackage{epsfig} 
\usepackage{amsmath}
\usepackage{amssymb}  
\usepackage[ruled, vlined, titlenotnumbered, linesnumbered]{algorithm2e}
\usepackage{multirow}
\usepackage{tabularx}
\usepackage{booktabs}
\usepackage{caption}

\usepackage[colorlinks, citecolor=green]{hyperref}

\usepackage{subcaption}
\usepackage{color,soul}
\usepackage[normalem]{ulem}

\hypersetup{
    colorlinks=true,
    linkcolor=blue,
    filecolor=magenta,      
    urlcolor=blue,
    pdftitle={Overleaf Example},
    pdfpagemode=FullScreen,
    }

\begin{document}

\title{\LARGE \bf
\textcolor[RGB]{81,131,236}{\textbf{A}}\textcolor[RGB]{216,81,64}{\textbf{I}}\textcolor[RGB]{241,191,66}{\textbf{S}}\textcolor[RGB]{81,131,236}{\textbf{P}}\textcolor[RGB]{88,166,91}{ \textbf{O}}:
Enhancing Depth Reliability for Robotic Manipulation of Non-Lambertian Objects via  \textcolor[RGB]{81,131,236}{\textbf{A}}ffine-\textcolor[RGB]{216,81,64}{\textbf{I}}nvariant \textcolor[RGB]{241,191,66}{\textbf{S}}hape \textcolor[RGB]{81,131,236}{\textbf{P}}ri\textcolor[RGB]{88,166,91}{\textbf{o}}r 
}

\author{
    Zhiming Chen$^{*1,6}$, 
    Linfang Zheng$^{*2}$, 
    Kun Zhang$^{6}$, 
    Hyung Jin Chang$^{5}$, 
    Wei Zhang$^{3,6}$, 
    Hongyu Yu$^{1}$, 
    Hua Chen$^{\dagger 4,6}$,
    
\thanks{Manuscript received: Decemember 18, 2025; Revised: February 27, 2026; Accepted: April 3, 2026. 
} 
\thanks{This paper was recommended for publication by Editor Markus Vincze upon evaluation of reviewers' comments. }
\thanks{$^*$\text{denotes equal contribution}, $^\dagger$\text{denotes} the corresponding author. }
\thanks{$^{1}$The Hong Kong University of Science \& Technology, $^{2}$The University of Hong Kong, $^{3}$Southern University of Science \& Technology, $^{4}$Zhejiang University, $^{5}$University of Birminghan, $^{6}$LimX Dynamics. \texttt{zhiming.chen@connect.ust.hk,
huachen@intl.zju.edu.cn}
}
\thanks{Digital Object Identifier (DOI): see top of this page.}

}

\markboth{ IEEE ROBOTICS AND AUTOMATION LETTERS. PREPRINT VERSION. ACCEPTED April, 2026}%
{CHEN \MakeLowercase{\textit{et al.}}: AISPO: Enhancing Depth Reliability for Robotic Manipulation of Non-Lambertian Objects via Affine-Invariant Shape Prior}


\maketitle

\begin{abstract}
Reliable depth perception is critical for robotic manipulation, especially for non-Lambertian objects such as transparent or highly specular surfaces, where raw depth measurements are often corrupted or missing. These failures frequently propagate to motion planning, resulting in invalid grasp poses and execution errors. We propose AISPO, a depth completion framework that improves depth reliability for manipulation in challenging sensing conditions. AISPO combines multi-scale RGB-D feature fusion with an affine-invariant shape prior to enforce geometric consistency and mitigate catastrophic depth failures. Unlike methods that focus primarily on average depth accuracy, our approach emphasizes physical plausibility and structural integrity of the predicted depth maps. Extensive benchmark evaluations demonstrate competitive performance and strong generalization to unseen objects and novel scenes. Real-world grasping experiments further show that enhanced depth reliability significantly improves manipulation success rates, particularly for transparent objects where many existing methods fail to produce physically usable depth estimates.
\end{abstract}


\section{INTRODUCTION}
\label{section:introduction}

Reliable depth perception is fundamental to robotic manipulation. 
Modern robotic systems~\cite{SynergiesBA}\cite{ManiSkill}\cite{IterativeVLN} heavily rely on depth sensors to estimate scene geometry for grasp planning, collision avoidance, and motion execution. 
However, when interacting with non-Lambertian objects—such as transparent containers, glassware, or highly specular metallic surfaces—raw depth measurements are often severely corrupted or entirely missing. 
In practical manipulation pipelines, such perception failures frequently propagate to motion planning, resulting in physically invalid grasp poses and execution failure. 
Therefore, improving depth reliability under these challenging conditions is critical for robust real-world robotic systems~\cite{AugmentedUnpairedData}\cite{SAID-NeRF}.

\begin{figure}[htbp]
    \centering
    \includegraphics[width=0.92\linewidth]{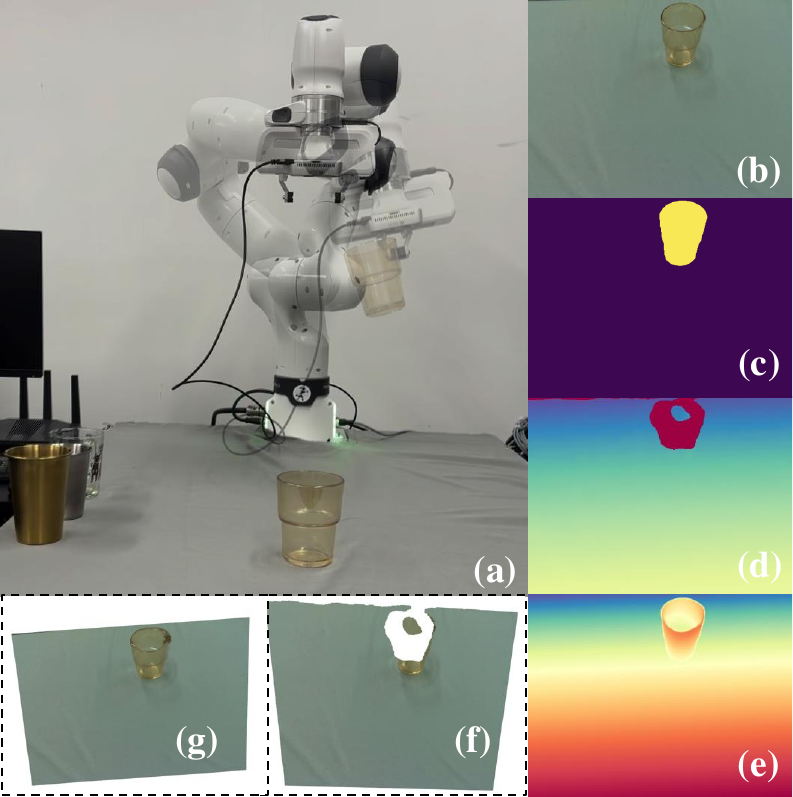}
    \caption{\textbf{Unlocking reliable depth sensing for robotic manipulation in the real world.} AISPO completes missing depth for challenging non-Lambertian objects—such as transparent and reflective items—by combining learned shape priors and multi-modal visual cues. Shown: (a) real-world robotic scene, (b) input RGB, (c) detected non-Lambertian object mask, (d) raw depth with large missing regions, (e) our completed depth, and the RGB-colored point clouds (f) and (g) generated from raw and completed depth respectively. Our approach enables significant improvement of the success rate on grasping non-lambertian objects.
    }
    \label{fig:franka}
    \vspace{-20pt}
\end{figure}

Existing depth completion methods aim to refine incomplete or noisy depth maps by leveraging RGB guidance or geometric constraints\cite{cleargrasp}\cite{LIDF}\cite{TransCG}. 
While significant progress has been achieved on public benchmarks, many approaches primarily optimize global depth error metrics (e.g., RMSE or MAE), which may not adequately reflect system-level robustness. 
In robotic manipulation, small average improvements are often less important than preventing catastrophic depth failures that produce distorted object geometry or hallucinated surfaces. 
Moreover, multi-view or optimization-based approaches~\cite{SAID-NeRF}\cite{ResidualNeRF}  can improve reconstruction fidelity but typically incur substantial computational overhead, limiting their suitability for time-sensitive robotic applications.

To address these challenges, we propose \textbf{AISPO}, a depth completion framework designed to enhance geometric consistency and reduce catastrophic estimation errors in manipulation-oriented scenarios. 
Our approach integrates multi-scale RGB-D feature fusion with an affine-invariant shape prior, enabling the network to better preserve structural integrity under non-Lambertian sensing conditions. 
By enforcing geometric regularity through a dedicated shape-prior autoencoder and a two-stage training paradigm, AISPO improves the physical plausibility of predicted depth maps rather than merely minimizing pixel-wise errors. We evaluate our method on public depth completion benchmarks to assess quantitative performance and generalization capability. 
More importantly, we validate its practical impact in real-world robotic grasping experiments. 
Our results demonstrate that reducing catastrophic depth artifacts directly translates into substantially higher grasp success rates, particularly for transparent objects where many existing methods fail to produce usable depth maps.

The main contributions of this work are summarized as follows:

\begin{enumerate}
    \item We propose AISPO, a depth completion framework that integrates multi-scale RGB-D feature fusion with an affine-invariant shape prior to enhance geometric consistency under non-Lambertian sensing conditions.
    
    \item We introduce a shape-prior autoencoder and a two-stage training strategy to enforce structural regularity and reduce catastrophic depth estimation failures.
    
    \item We demonstrate through extensive benchmark evaluation and real-world robotic experiments that improved depth reliability significantly enhances manipulation performance, especially in challenging transparent object scenarios.
\end{enumerate}
\section{RELATED WORK} \label{section:related-work}


\textbf{Single-view RGB-D Depth Completion:} ClearGrasp \cite{cleargrasp} pioneered the use of single-view RGB-D input for transparent object depth completion, employing three separate models to predict the mask, surface normal, and occlusion boundary, followed by a global optimization step to obtain the final depth. Zhu et al. \cite{LIDF} proposed that the depth of a transparent object can be inferred from nearby opaque regions using a trained local implicit depth function (LIDF). TransparentNet \cite{transparent_net} lifts depth maps to point clouds for completion. DFNet \cite{TransCG} utilizes a U-Net architecture to encode inaccurate depth maps at multiple resolutions. More recent works, such as SwinDRNet \cite{dreds} and TODE-Trans \cite{TODE_Trans}, leverage the Swin Transformer \cite{swin_transformer} to better capture global context. However, the challenge of accurately handling non-Lambertian surfaces remains largely unresolved.

\textbf{Monocular Depth Estimation:} Recently, monocular depth estimation has attracted a lot of research attention. However, most of them \cite{DA2}\cite{MiDaS}\cite{Marigold} focus on relative depth estimation, which is more beneficial for AIGC applications \cite{zest}\cite{RealmDreamer} and does not directly apply to real-world robotic manipulations. Absolute depth estimation methods \cite{Metric3D_v2} exhibit significant deviations from the ground truth for diffuse objects - let alone challenging cases involving transparent or specular surfaces. PromptDA \cite{promptDA} uses LiDAR as the prompt to guide the Depth Anything model \cite{DA2} for accurate metric depth output, which has the potential to generalize with transparent and specular objects. Depth-Pro~\cite{depth-pro} is a foundation model trained on various public datasets. MODEST \cite{MODEST} proposes a monocular framework to jointly predict depth and segmentation for transparent objects where a semantic and geometric fusion module that can better leverage the complementary information of both tasks. Nevertheless, these methods have limited performance in depth accuracy due to the ill-posed nature of the monocular depth estimation problem lacking direct metric input. Pi3~\cite{pi3} and Depth AnythingV3~\cite{DA3} are newer 3D vision foundation models which can also apply for single view case.

\textbf{Multi-view RGB Depth Prediction:} Multi-view methods typically leverage observations from multiple perspectives to benefit from richer visual information. In this work, we treat stereo approaches as a special case of two-view methods within the multi-view category. SimNet \cite{SimNet} explored a single multi-headed neural network as a multi-task framework using stereo data as input, utilizing multiple visual perceptions—including segmentation masks, 3D oriented bounding boxes, object keypoints, and disparity as output to support transparent object manipulation. D3RoMa \cite{d3roma} leverages stereo image pairs to predict the depth of non-Lambertian objects using a denoising diffusion probabilistic model \cite{DDPM}; however, its non-negligible training time and demand for large amounts of data limit its scalability. MVTrans \cite{MVTrans} extends stereo methods by introducing an end-to-end multi-view architecture with multiple perceptual capabilities, thereby avoiding reliance on unreliable depth maps from RGB-D sensors. Nevertheless, multi-view methods~\cite{ClearDepth}~\cite{StereoAnything} still struggle with high computational costs and latency.

\textbf{NeRF-based Methods:} Building upon recent advances in neural radiance fields (NeRF) \cite{NeRF}, DexNeRF \cite{DexNeRF} became the first to extend multi-view depth estimation by employing an implicit neural radiance field to represent transparent objects, although its optimization process required several hours. EvoNeRF \cite{EvoNeRF} improves upon DexNeRF by integrating Instant-NGP \cite{InstantNGP}, a faster variant of NeRF. Instead of training on a fixed set of images, EvoNeRF incrementally optimizes from a stream of images captured during robot motion. However, it still relies on dense input views and requires scene-specific training prior to each grasping attempt. GraspNeRF \cite{Grasp-NeRF} later accelerated inference by leveraging a generalizable NeRF framework. Methods such as SAID-NeRF \cite{SAID-NeRF} and ResidualNeRF \cite{ResidualNeRF} further improve robustness by decoupling the background. Despite these advances, existing NeRF-based approaches often struggle with complex light interactions, particularly for highly transparent objects (e.g., wine glasses) or specular surfaces (e.g., kitchen foil), where the lack of stable visual features and strong view-dependent appearance variations pose significant challenges.

\section{METHOD} \label{section:method} 

\subsection{Problem Setup}
This section presents our depth-sensing framework designed to address the severe depth corruption commonly observed in non-Lambertian objects—such as transparent and specular surfaces—during robotic manipulation. The framework aims to enhance the accuracy of dense depth estimation, particularly for these challenging yet ubiquitous object types. Given an RGB image \( I \in \mathbb{R}^{3 \times H \times W} \) capturing color and texture information, and a raw, often corrupted depth image \( D_{raw} \in \mathbb{R}^{H \times W} \) providing geometric cues, our framework learns a mapping \( f_\theta(\cdot) \) to predict the corresponding restored depth image \( D_\text{restored} \):

\begin{equation}  
D_{\text{restored}} = f_{\theta}(I, D_{raw})  
\end{equation}

\subsection{Affine-Invariant Shape Prior} \label{subsection:dpa}

\begin{figure}[htbp]
    \centering
    \includegraphics[width=0.95\linewidth]{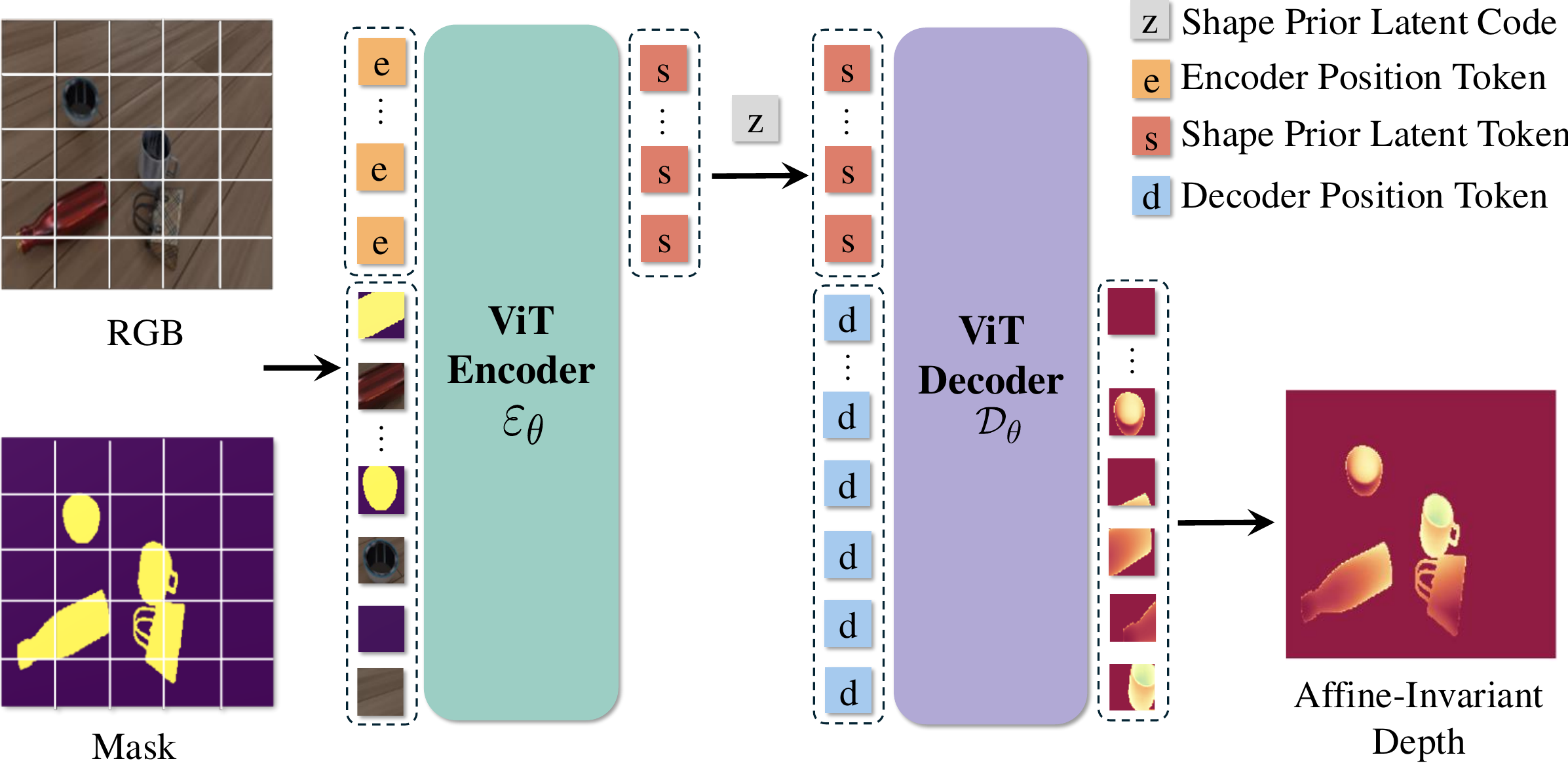}
   \caption{\textbf{Overview of the shape prior auto-encoder architecture.}
     Encoder $\varepsilon_\theta$ processes the RGB image and the non-Lambertian object mask to extract the shape prior latent code $z$. Subsequently, decoder $\mathcal{D}_\theta$ decodes $z$ into an affine-invariant shape representation of the non-Lambertian object. Notably, the output contains only the object-level shape prior, with background information fully excluded.
    }
    \label{fig:dpa}
    \vspace{-15pt}
\end{figure}

As illustrated in Fig.~\ref{fig:dpa}, our depth completion framework for non-Lambertian objects starts by estimating an affine-invariant shape prior. Given an RGB image $I$ and an object mask $M$ typically obtained via a foundation model like Grounded-SAM2~\cite{Grounded-SAM2}—we train an autoencoder to reconstruct a masked, affine-invariant depth map. The input images are first encoded by a Vision Transformer (ViT) encoder  $\varepsilon_\theta$ into a latent shape code $z$, which is subsequently decoded to produce the corresponding affine-invariant shape representation.

For the subsequent depth completion stage, we freeze the encoder weights and discard the decoder $\mathcal{D}_\theta$, retaining only the encoder to extract the shape prior latent code. To generate the ground-truth depth map, we first mask out the background and then normalize the depth values within each object instance, resulting in an affine-invariant object shape representation $D_{\text{shape}}$. This normalization eliminates unknown scale and shift variations across objects. The encoder $\varepsilon_\theta$ and decoder $\mathcal{D}_\theta$ are optimized using an $L_1$ loss.

\begin{equation}
    \mathcal{L}_\text{full} = \frac{1}{N}\sum_{i=1}^{N}\|D_\text{pred}^{i} -  D_{\text{shape}}^{i} \|_{1}
\end{equation}

where $D_\text{pred}^{i}$ represents the scaled and shifted depth prediction. To further emphasize the object region, we also incorporate a masked loss term.
  
\begin{equation}
     \mathcal{L}_\text{masked} = \frac{1}{N}\sum_{i=1}^{N}\|D_\text{pred}^{i}\odot M_i - D_\text{shape}^{i} \odot M_i \|_{1}
\end{equation}

where $\odot$ denotes element-wise masking using the object mask $M_i$. Additionally, we incorporate a Sobel~\cite{sobel} gradient loss to enhance training stability and sharpen object shape boundaries.

\begin{equation}
     \mathcal{L}_\text{Sobel} = \frac{1}{N}\sum_{i=1}^{N}\|G(D_\text{pred}^{i}) - G(D_{\text{shape}}^{i})\|_{1}^2
\end{equation}
where $G(D)=\sqrt{G_x * D + G_y * D}$, and
\begin{equation}
G_x = \begin{bmatrix}
-1 & 0 & 1 \\
-2 & 0 & 2 \\
-1 & 0 & 1
\end{bmatrix}, \quad
G_y = \begin{bmatrix}
-1 & -2 & -1 \\
0 & 0 & 0 \\
1 & 2 & 1
\end{bmatrix}
\end{equation}

The total training objective for the shape prior autoencoder combines the aforementioned components into the following loss function:

\begin{equation}
    \mathcal{L}_\text{shape}=\alpha_{\text{full}} \cdot \mathcal{L}_\text{full}+\alpha_\text{masked} \cdot \mathcal{L}_\text{masked}+ \alpha_\text{Sobel} \cdot \mathcal{L}_\text{Sobel}
\end{equation}

\subsection{Overall Framework}

\begin{figure*}
    \centering
    \includegraphics[width=0.92\linewidth]{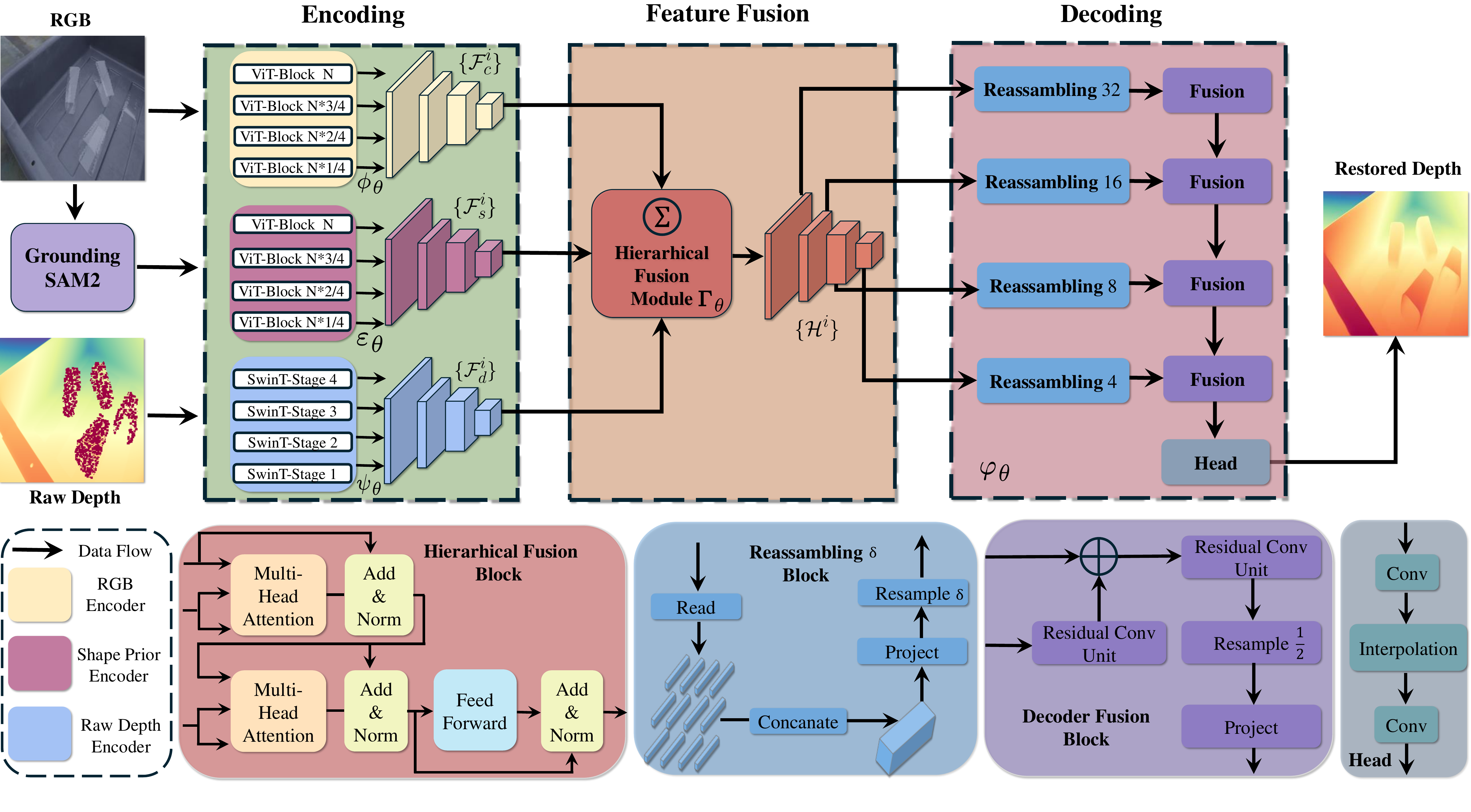}
    \caption{ 
    \textbf{Framework Overview.} Our framework extracts complementary cues via three parallel encoders. The RGB encoder $\phi_\theta$ processes image $I$ for color and texture. Concurrently, a pretrained shape‑prior encoder $\varepsilon_\theta$ takes $I$ and a non‑Lambertian mask (from Grounded‑SAM2~\cite{Grounded-SAM2}) to capture object‑shape features. A Swin‑Transformer‑based depth encoder $\psi_\theta$ handles the raw depth input. These streams yield multi‑scale intermediate features  $\{ \mathcal{F}_c^i \}, \{ \mathcal{F}_s^i \}, \{ \mathcal{F}_d^i \}$, which are fused through cross‑attention in a hierarchical fusion module $\Gamma_\theta$. The fused representation is iteratively refined and decoded by $\varphi_\theta$ to produce the final restored depth $D_\text{restored}$.
    }
    
    \label{fig:framework}
    \vspace{-10pt}
\end{figure*}

\subsubsection{Feature Extraction}
As illustrated in Fig.~\ref{fig:framework}, our framework employs three dedicated encoders to extract complementary cues from RGB and raw depth inputs, each designed to address specific challenges in scene understanding and depth completion. First, an RGB encoder $\phi_{\theta}$, based on the vision foundation model DINOv2 \cite{DINOv2}, extracts color and texture features ${\mathcal{F}_c^i}$ from the RGB input. While this encoder effectively leverages rich photometric information to distinguish objects with similar geometry but distinct appearances, RGB data alone lack explicit depth cues and are therefore inadequate for accurate metric depth estimation. To address this limitation, we introduce a raw depth encoder $\psi_{\theta}$, based on a Swin-Transformer architecture \cite{swin_transformer}, to extract geometric features ${\mathcal{F}_d^i}$ from raw depth inputs. The hierarchical design and shifted window mechanism of the Swin-Transformer enable effective modeling of both long-range dependencies and local spatial details, making it well-suited for processing challenging surfaces such as transparent or specular materials \cite{TODE_Trans}. This design achieves an effective balance between local precision and global context, which is essential for robust depth estimation.

However, raw depth images are often corrupted or contain missing pixels due to sensor noise, reflections, or transparency. To mitigate these issues, we explicitly incorporate a shape prior encoder $\varepsilon_{\theta}$, introduced in Sec.~\ref{subsection:dpa}, which extracts object-level shape features from the RGB image and its corresponding object mask. By leveraging learned shape priors, this encoder compensates for incomplete or noisy depth data, thereby enhancing the overall robustness of the system.
By integrating these three complementary encoders, our framework effectively fuses color cues, geometric features, and shape priors to build a comprehensive scene representation, thereby addressing the inherent limitations of each individual modality. From each encoder, we extract a set of multi-scale features, specifically from four intermediate stages of their blocks:

\begin{align}
    \{\mathcal{F}_c^i\}_{i=0,1,2,3} &= \phi_{\theta}({I}) \\
    \{\mathcal{F}_s^i\}_{i=0,1,2,3} &= \varepsilon_{\theta}(I, \text{\small{GroundedSAM2}}(I)) \\
    \{\mathcal{F}_d^i\}_{i=0,1,2,3} &= \psi_{\theta}({D_{raw}})
\end{align}

\subsubsection{Cross-Attention based Feature Fusion}
Given the intermediate features $\{\mathcal{F}_c^i\}, \{\mathcal{F}_d^i\}, \{\mathcal{F}_s^i\}$ extracted from the three encoders, our fusion module $\Gamma_\theta$ employs cross-attention transformers to integrate them into a fused hierarchical representation: $\mathcal{H}^i=\Gamma_\theta(\mathcal{F}_c^i, \mathcal{F}_d^i, \mathcal{F}_s^i)$. Before fusion, all intermediate features are projected to the same dimension. The fusion proceeds through sequential cross-attention operations. Given  $Q_{s} = \mathcal{F}_s^i\cdot W_q,K_c=\mathcal{F}_c^i\cdot W_k,V_c=\mathcal{F}_c^i\cdot W_v$, we first obtain a fused feature $H_{\mathcal{F}_s^i \rightarrow \mathcal{F}_c^i}$ from the cross attention operation between $\mathcal{F}_s^i$ and $\mathcal{F}_c^i$. That is:
\begin{equation}
    \begin{aligned}
        &H_{\mathcal{F}_s^i \rightarrow \mathcal{F}_c^i} = \text{Softmax} \left( \frac{Q_s \cdot K_c^T}{\sqrt{d_K}} \right) \cdot V_c \\
        &H_{\mathcal{F}_s^i \rightarrow \mathcal{F}_c^i}=\text{LayerNorm}(H_{\mathcal{F}_s^i \rightarrow \mathcal{F}_c^i}+\mathcal{F}_s^i)
    \end{aligned}
\end{equation}

Next, we apply cross-attention between $H_{\mathcal{F}_s^i \rightarrow \mathcal{F}_c^i}$ and $\mathcal{F}_d^i$ with $Q_{sc} = H_{\mathcal{F}_s^i \rightarrow \mathcal{F}_c^i} \cdot W_q^\prime, K_d = \mathcal{F}_d^i \cdot W_k^\prime, V_d = \mathcal{F}_d^i \cdot W_v^\prime$. 
\begin{equation}
    \begin{aligned}        
    &\mathcal{H}_{\mathcal{F}_s^i \rightarrow \mathcal{F}_c^i \rightarrow \mathcal{F}_d^i} = \text{Softmax} \left( \frac{Q_{sc} \cdot K_{d}^T}{\sqrt{d_K}} \right) \cdot V_d \\
    &\mathcal{H}_{\mathcal{F}_s^i \rightarrow \mathcal{F}_c^i \rightarrow \mathcal{F}_d^i} = \text{LayerNorm}(H_{\mathcal{F}_s^i \rightarrow \mathcal{F}_c^i \rightarrow \mathcal{F}_d^i} + \mathcal{F}_d^i)
    \end{aligned}
\end{equation}

Finally, a two-layer feedforward network refines the fused representation:

\begin{equation}
\begin{aligned}
    &\mathcal{H}^i = \text{FFN}(H_{\mathcal{F}_s^i \rightarrow \mathcal{F}_c^i \rightarrow \mathcal{F}_d^i}) \\
    &\mathcal{H}^i = \text{LayerNorm}(\mathcal{H}^i + \mathcal{H}_{\mathcal{F}_s^i \rightarrow \mathcal{F}_c^i \rightarrow \mathcal{F}_d^i})
\end{aligned}
\end{equation}


\subsubsection{Dense Depth Prediction}
Following MiDaS~\cite{MiDaS} and DepthAnythingV2~\cite{depth_anything_v2}, we use the DPT~\cite{DPT} decoder architecture in the decoder $\varphi_\theta$ for depth regression. The fused multi-scale hierarchical feature tokens are assembled into image-like representations at various resolutions, which are then progressively refined using RefineNet-based fusion blocks~\cite{RefineNet}\cite{Xian2018MonocularRD_refinenet}. These blocks integrate low- and high-resolution features through residual convolutional units and iterative fusion mechanisms. Finally, a convolutional layer generates the dense depth prediction:

\begin{equation}
\mathrm{D_{\text{restored}}} = \varphi_{\theta}(\{\mathcal{H}^i\}_{i=0,1,2,3})    
\end{equation}

\subsubsection{Optimization}
During training, we freeze the parameters of the RGB encoder $\phi_\theta$ and the pre-trained shape prior encoder $\varepsilon_\theta$. The remaining modules ${ \psi_\theta, \Gamma_\theta, \varphi_\theta }$ are optimized using a combination of full and masked loss:

\begin{equation}
    \mathcal{L} =\beta_{\text{full}} \cdot\mathcal{L}_{\text{full}} + \beta_{\text{masked}} \cdot \mathcal{L}_{\text{masked}} \\
\end{equation}
Where $\mathcal{L}_{\text{full}} = \frac{1}{N}\sum_{i=1}^{N}\| D_{\text{restored}}^{i} - D_{\text{gt}}^{i}\|_{1}$, $\mathcal{L}_{\text{masked}} = \frac{1}{N}\sum_{i=1}^{N}\| D_{\text{restored}}^{i}  \odot M_i - D_{\text{gt}}^{i} \odot M_i \|_{1}$, and $\odot$ denotes the mask operator.

\begin{figure*}[t]
    \centering
    \includegraphics[width=0.92\linewidth]{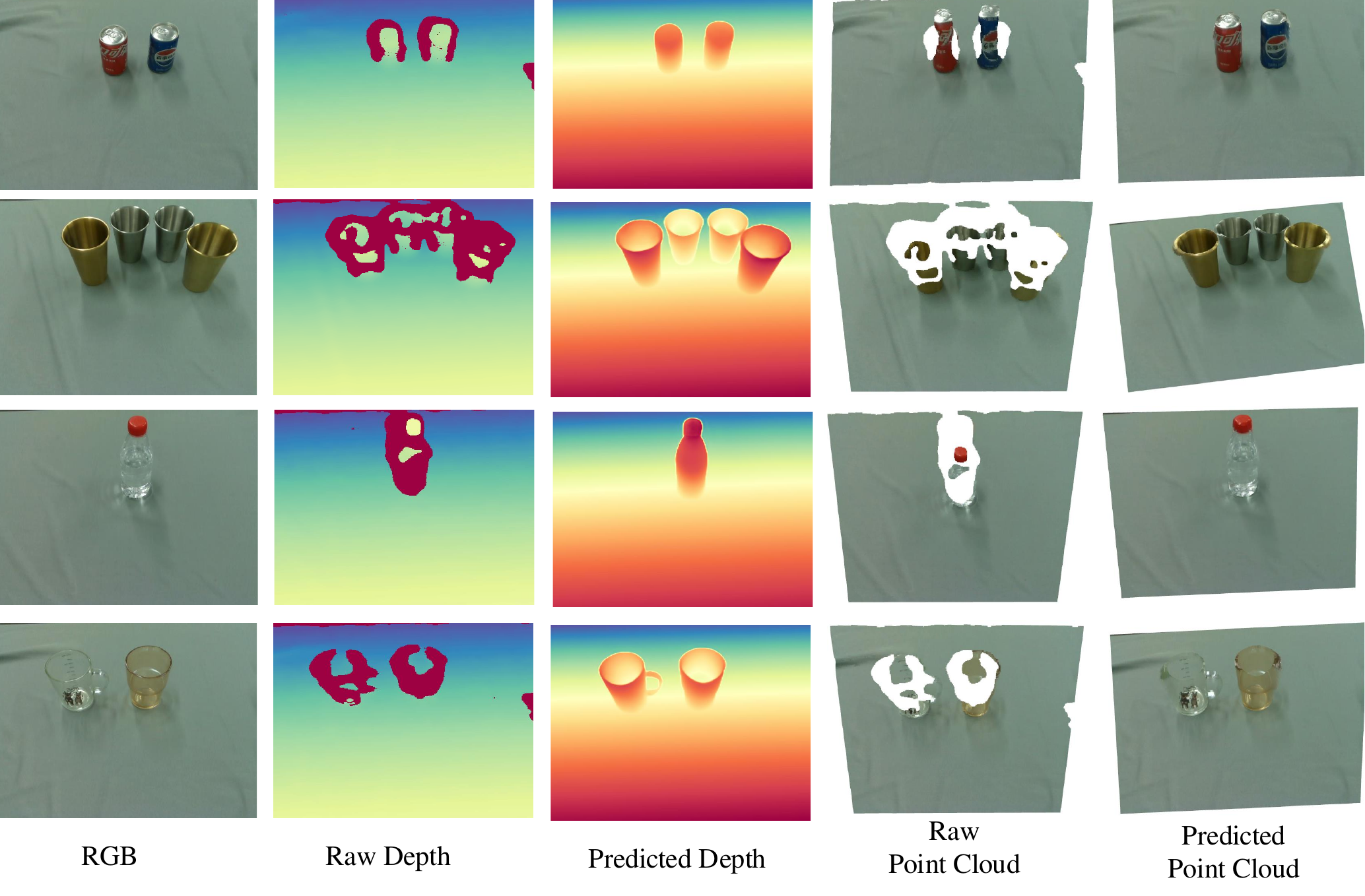}
      \caption{\textbf{Real‑world qualitative results on non‑Lambertian objects in robotic manipulation environment.} Using only synthetic data (DREDS‑CatKnown~\cite{dreds}) for training the framework, our model generalizes to real household objects with severe depth corruption (e.g., reflective cans, water‑filled bottles, transparent cups). It recovers dense depth and complete 3D geometry, accurately reconstructing fine structures like thin handles, demonstrating strong zero‑shot robustness.}
    \label{fig:real-world-result}
    \vspace{-15pt}
\end{figure*}

\section{Experiment} \label{sec:experiment}


\subsection{Experiment Setup}
Our framework adopts a two-stage training paradigm. In the first stage, we train a Vision Transformer (ViT)-based shape prior autoencoder on the DREDS-CatKnown dataset~\cite{dreds}, which comprises 100,200 training and 19,380 testing RGB-D images across 1,801 ShapeNetCore \cite{ShapeNet} objects rendered with randomized specular, transparent, and diffuse textures. Once trained, this module remains fixed in all subsequent experiments.

Both the encoder $\varepsilon_\theta$ and decoder $\mathcal{D}_\theta$ are implemented with $8$ ViT blocks, employing a patch size of $16 \times 16$, $8$ attention heads, and a feature dimension of $768$. The model is trained for $300$ epochs on eight NVIDIA A100 GPUs with a batch size of $128$, using the AdamW optimizer at a learning rate of $8.0 \times 10^{-5}$. The loss weighting coefficients are set to $\alpha_{\text{full}}=1.0$, $\alpha_{\text{masked}}=0.8$, and $\alpha_{\text{Sobel}}=0.5$. After pre-training, only the encoder $\varepsilon_\theta$ is retained for the second stage.

In the second stage, we train the full model on the DREDS‑CatKnown dataset for quality comparison. The training runs for 300 epochs with a batch size of 64, using the AdamW optimizer (learning rate $8.0 \times 10^{-5}$) and loss weights $\beta_{\text{full}} = 1.0$, $\beta_{\text{masked}} = 0.8$. The RGB encoder $\phi_\theta(\cdot)$ is initialized with the pretrained DINOv2‑base model (Hugging Face). The raw‑depth encoder $\psi_\theta(\cdot)$ is a Swin‑Transformer with patch size 4, a single input channel, an embedding dimension of 128, and Swin-Transformer blocks arranged as (2, 2, 18, 2) across its four stages. The number of attention heads per stage is (4, 8, 16, 32), with a window size of 16. The hierarchical fusion module $\Gamma_\theta(\cdot)$ employs four cross-attention blocks to process intermediate features at different scales. It uses an embedding dimension of $1024$ and $8$ attention heads. The decoder $\varphi_\theta(\cdot)$ takes an input channel of $192$; its four fusion branches have output channels of $256$, $512$, $1024$, and $1024$, respectively. All models are trained and evaluated at a resolution of $256\times256$. For real-world robotic grasping, predictions are interpolated to the original input resolution.

We compare our method against four state-of-the-art approaches. SwinDRNet~\cite{dreds} and DFNet~\cite{TransCG} serve as baselines for the DREDS-CatKnown and TransCG datasets, respectively. DepthAnythingV2 (DA2)~\cite{DA2} is a foundation model of single view depth estimation for comparison. PromptDA~\cite{promptDA} is a recent monocular depth estimation model that fine-tunes the previous foundation model~\cite{DA2}. Since the related field is evolving quite fast, we further add Pi3~\cite{pi3}, and  another two depth estimation foundation models: DepthAnythingV3(DA3$^*$\footnotemark)~\cite{DA3} and Depth-Pro$^*$\footnotemark[1]~\cite{depth-pro}.

\footnotetext[1]{$*$ indicates results reproduced with publicly available weights, since these methods were trained on diverse datasets and only inference code was released.}

\subsection{Robotic Manipulation Experiment}
As shown in Fig.~\ref{fig:franka}, we evaluate our method in real-world grasping experiments using a Franka robotic arm, with all models running on an NVIDIA RTX 3090 GPU. Our depth completion model for manipulation is trained on aforementioned synthetic DREDS~\cite{dreds} dataset. To ensure fairness, all methods share the same grasp motion planning configuration \cite{moveit2025}, differing only in the input depth maps. The grasp point will first extracted from the depth map, then a feasible trajectory will be planned to execute the action. For statistical reliability, we conduct 30 independent trials per method on specular objects and 45 trials on transparent objects, reflecting the higher difficulty of the latter. The success rates are reported in Table~\ref{tab:grasp-experiment}. On specular objects, baseline methods achieve non-zero success rates. However, on transparent objects, baseline methods consistently fail due to inaccurate depth predictions that result in invalid grasp proposals. In contrast, our method substantially improves grasp success rates in both settings.

\begin{table}[hbtp]
    \tabcolsep = 2.0pt
    \centering
    \renewcommand{\arraystretch}{1.2}
    \begin{tabular}{l|cccc}
        \toprule
        \text{Method} & \text{Specular SR} & \text{Transparent SR} &  \text{Total SR} \\
        \midrule
        Raw & 0.13 (4/30) & 0 (0/45) & 0.05  \\
        SwinDRNet & 0.17 (5/30) & 0 (0/45) & 0.07 \\
        DFNet & \underline{0.63} (19/30) & \underline{0} (0/45)  & \underline{0.25} \\
        Ours & \textbf{1.00} (30/30) & \textbf{0.89} (40/45) & \textbf{0.93} \\
        \bottomrule
    \end{tabular}
    \caption{Tabletop grasping success rates (SR) on specular and transparent objects. Transparent objects represent a challenging perception regime where inaccurate depth estimation directly leads to invalid grasp planning. }
    \label{tab:grasp-experiment}
    \vspace{-15pt}
\end{table}

\subsection{Failure Analysis / Real-World Qualitative Results}

Fig.~\ref{fig:real-world-result} shows qualitative results from robotic manipulation environment on real-world non-Lambertian objects, including coke cans, metal cups, water-filled bottles, and transparent mugs. Our analysis identifies two main categories of failure:

\begin{itemize}
    \item \textbf{Upstream Perception Failures} -- Errors from early stages, such as inaccurate instance segmentation under overlapping or occluded objects, transparent or reflective surfaces, and small or low-contrast items, can propagate to depth reconstruction and grasp planning. These failures occasionally produce incomplete object masks or distorted point clouds.

    \item \textbf{Depth Estimation Failures} -- Localized mismatches between the predicted affine-invariant shape prior and actual object geometry may occur, for example, due to complex refraction in transparent objects like glasses or thin structures like handles. These distortions are usually minor and rarely prevent successful grasps on the main object body.
\end{itemize}

Trained solely on the synthetic DREDS-CatKnown dataset~\cite{dreds},  AISPO generates dense, geometrically coherent depth maps that recover fine structures accurately. This enables valid grasp proposals even in challenging scenarios where baseline methods fail. By explicitly distinguishing failure modes, we provide clearer insight into system behavior and directions for future improvement, while demonstrating that reducing catastrophic depth failures directly improves real-world grasp success.

\begin{figure*}[hbpt]
    \centering
    \includegraphics[width=0.95\linewidth]{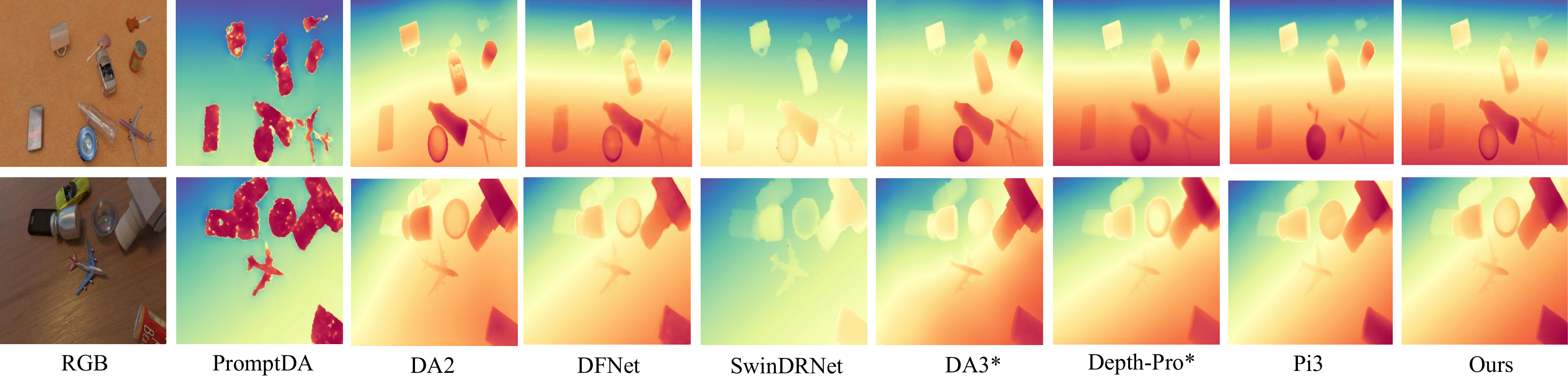}
    \caption{ \textbf{Qualitative Comparison on the DREDS-CatKnown dataset \cite{dreds}.} Each column (from left to right) shows the RGB image, the predictions from the baselines and our method.}
    \label{fig:dreds_result}
    \vspace{-10pt}
\end{figure*}

\begin{figure*}[hbpt]
    \centering
    \includegraphics[width=0.95\linewidth]{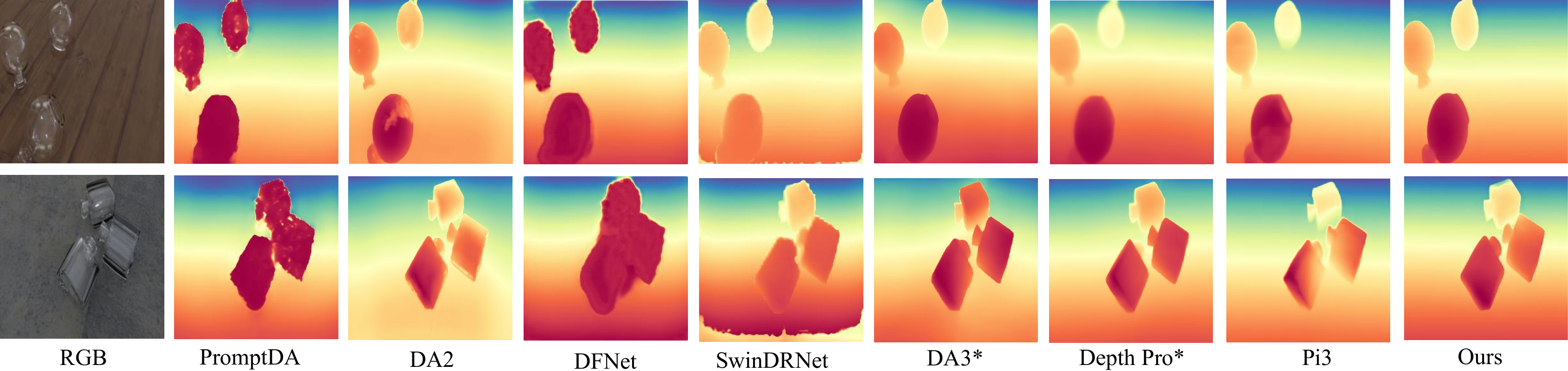}
    \caption{\textbf{Zeroshot qualitative comparison on the ClearGrasp dataset~\cite{cleargrasp}.} Models trained on the DREDS-CatKnown dataset \cite{dreds} are directly evaluated on the ClearGrasp dataset.
    }
    \label{fig:cleargrasp_result}
    \vspace{-15pt}
\end{figure*}

\subsection{Benchmark Evaluation: Depth Completion Quality}

\subsubsection{Quality Evaluation}


We evaluate our method and baselines on the synthetic DREDS-CatKnown dataset~\cite{dreds}, with qualitative results shown in Fig.~\ref{fig:dreds_result}. Our approach effectively preserves object geometry and depth consistency, achieving performance on par with state-of-the-art methods. 


To further assess generalization ability, we conduct zero-shot evaluations on the synthetic ClearGrasp dataset~\cite{cleargrasp}. As shown in Fig.~\ref{fig:cleargrasp_result}, our method adapts well to unseen transparent objects and successfully recovers fine geometric details over the baselines, demonstrating strong generalization capability.

\subsubsection{Quantitative Evaluation}

\begin{figure}[htbp]
    \centering
    \includegraphics[width=0.90\linewidth]{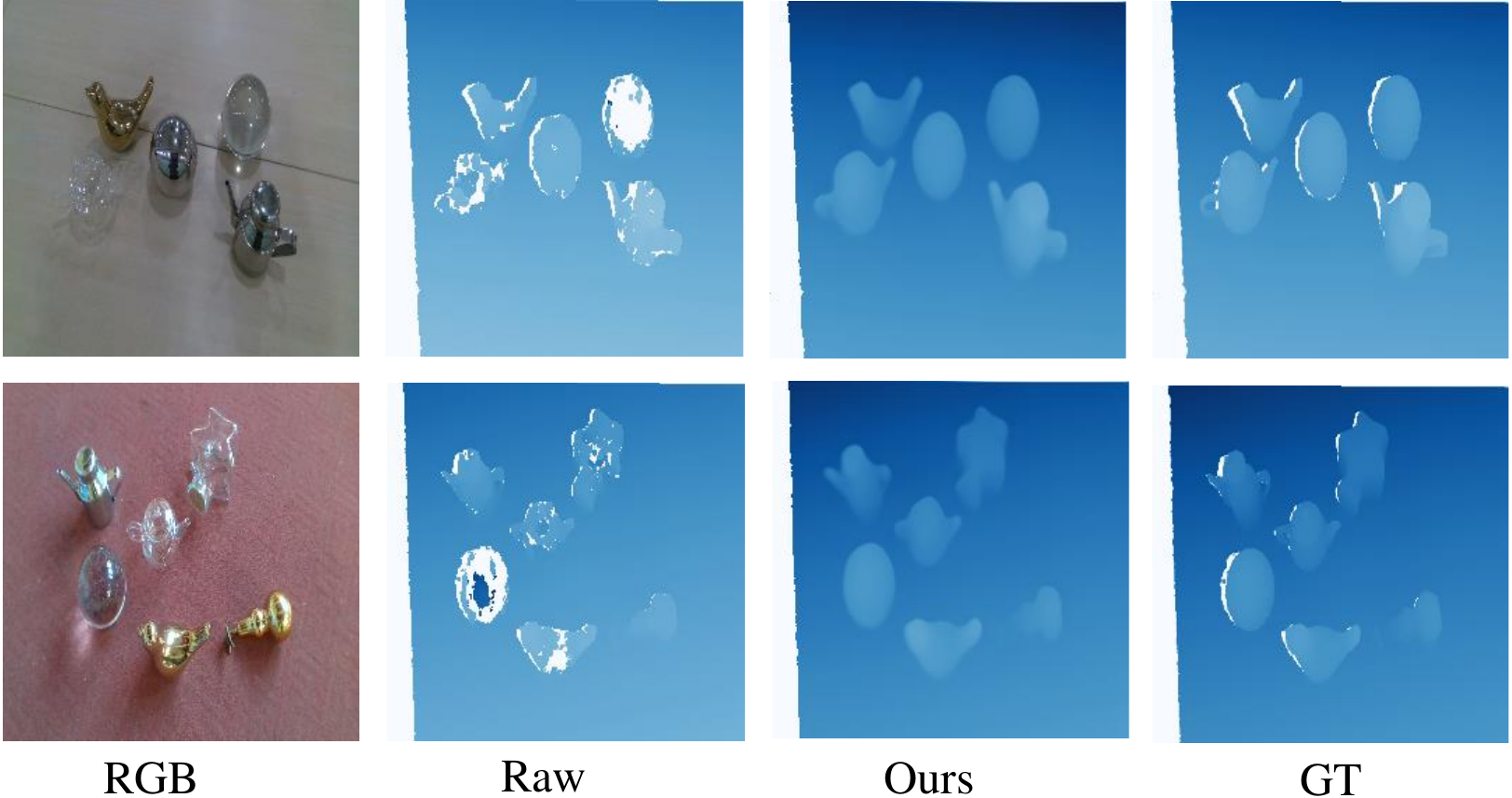}
    \caption{\textbf{Qualitative results of our method on the STD-CatNovel dataset~\cite{dreds}}.  For better visualization, we crop our prediction to align with the ground-truth missing region.}
    \label{fig:std_result}
    \vspace{-10pt}
\end{figure}

\begin{table}[htbp]
    \tabcolsep = 2.0pt
    \centering
    \renewcommand{\arraystretch}{1.2}
    \begin{tabular}{l|cccccc}
        \toprule
        \text{Method} & \text{RMSE}$\downarrow$ & \text{MAE}$\downarrow$ & \text{REL}$\downarrow$ & \text{$\delta_{1.05}$}$\uparrow$ & \text{$\delta_{1.10}$}$\uparrow$ & \text{$\delta_{1.25}$}$\uparrow$ \\
        
        \midrule
        \midrule
        PromptDA & 0.203 & 0.128 & 0.281 & 31.23 & 50.19 & 67.32 \\
        DA2 & 0.035 & 0.017 & 0.074 & 85.02 & 94.19 & 98.56 \\
        TransCG & 0.048 & 0.033 & 0.095 & 51.35 & 79.73 & 98.75 \\
        SwinDRNet & \underline{0.031} & \underline{0.012} & \textbf{0.046} & {93.98} & {97.69} & {98.95} \\
        DA3$^*$ & 0.150 & 0.144 & 0.291 & 8.64 & 15.76 & 33.76 \\
        Depth-Pro$^*$ & 0.198 & 0.191 & 0.431  & 8.32 & 14.77  & 36.10 \\
        Pi3 & 0.034 & \textbf{0.007} & 0.056 & \underline{96.46} & \underline{97.87} & \underline{99.06} \\
        Ours & \textbf{0.030} & \textbf{0.007} & \underline{0.047} & \textbf{97.26} & \textbf{98.59} & \textbf{99.19}\\
        \bottomrule
    \end{tabular}
    \caption{Quantitative results on the STD-Novel dataset. }
    \label{tab:STD-CatNovel_result}
    \vspace{-10pt}
\end{table}



\begin{figure}[htbp]
    \centering
    \includegraphics[width=0.90\linewidth]{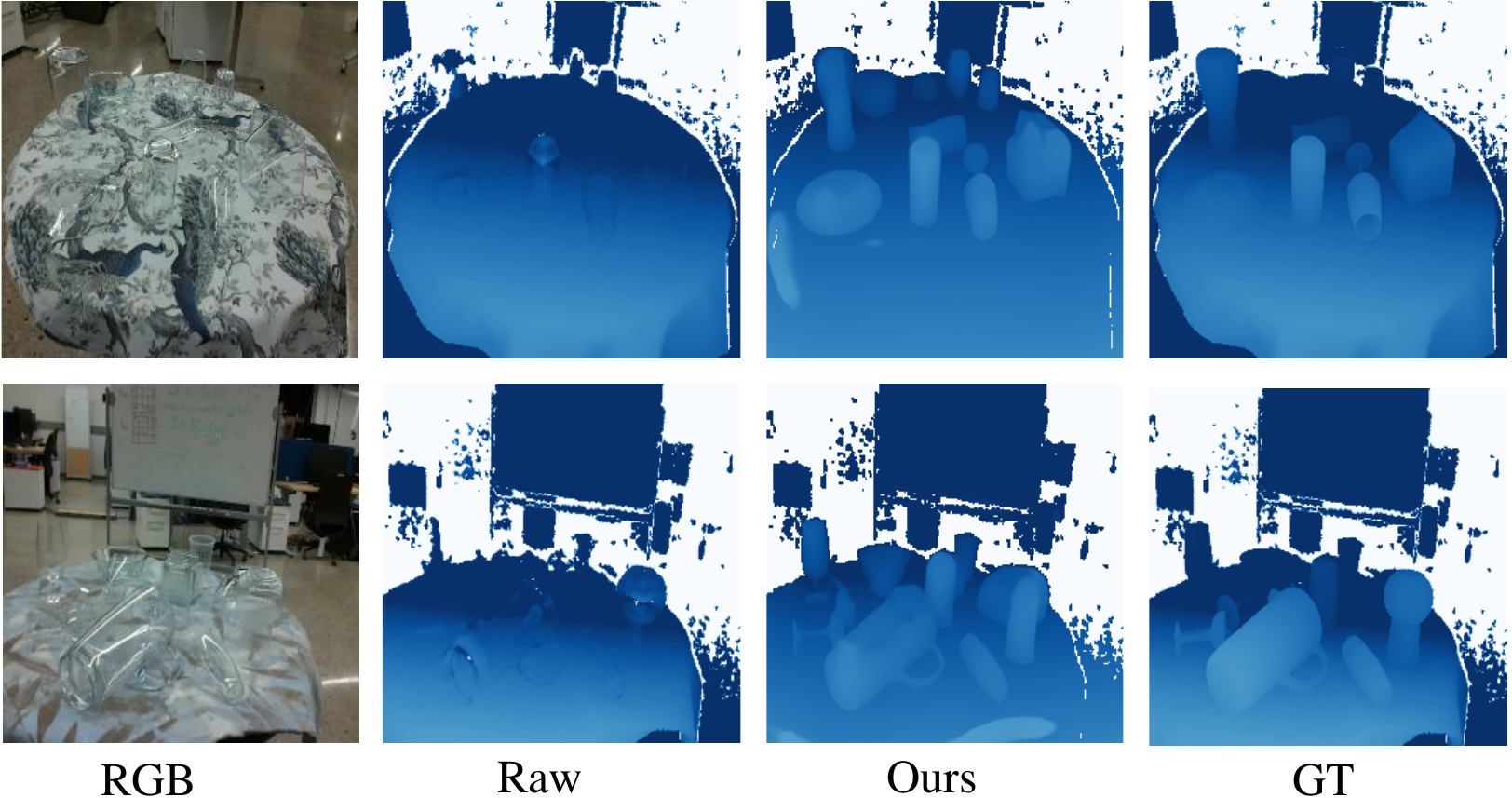}
    \caption{
    \textbf{Qualitative results on ClearPose~\cite{clearpose}.} Columns show RGB input, raw depth, our prediction, and ground truth. 
    }
    \label{fig:clearpose_result}
    \vspace{-20pt}
\end{figure}

We evaluate depth completion on two real-world datasets, STD-CatNovel~\cite{dreds} and ClearPose~\cite{clearpose}. STD-CatNovel contains novel object categories in diverse scenes, including transparent and specular objects, while ClearPose features challenging scenarios with transparent and translucent objects under occlusion and non-planar surfaces. As shown in Fig.~\ref{fig:std_result}, Fig.~\ref{fig:clearpose_result}, Table~\ref{tab:STD-CatNovel_result}, and Table~\ref{tab:clearpose_result}, our method effectively refines raw depth and reconstructs object structures. It achieves the best performance on STD-CatNovel and ranks second on ClearPose, with only marginal gaps to the top method.

\begin{table}[htbp]
    \tabcolsep = 2.0pt
    \centering
    \renewcommand{\arraystretch}{1.2}
    \begin{tabular}{l|cccccc}
        \toprule
        \text{Method} & \text{RMSE}$\downarrow$ & \text{MAE}$\downarrow$ & \text{REL}$\downarrow$ & \text{$\delta_{1.05}$}$\uparrow$ & \text{$\delta_{1.10}$}$\uparrow$ & \text{$\delta_{1.25}$}$\uparrow$ \\
        
        \midrule
        \midrule
        PromptDA & 0.243 & 0.133 & 0.175 & 37.70 & 53.96 & 80.16 \\
        DA2 & \underline{0.070} & {0.038} & {0.065} & {68.85} & {86.78} & 97.22 \\
        TransCG & 0.077 & 0.047 & 0.080 & 57.58 & 79.93 & 97.06 \\
        SwinDRNet & 0.081 & 0.048 & 0.073 & 52.41 & 79.72 & {97.66} \\
        DA3$^*$ & 0.382 & 0.360 & 0.453 & 0.29 & 0.57 & 2.77 \\
        Depth-Pro$^*$ & 0.218 & 0.132 & 0.217 & 23.99 & 45.36 & 81.58 \\
        Pi3 & \textbf{0.057} &\textbf{ 0.027} & \textbf{0.050} & \textbf{83.03} & \textbf{93.76} & \textbf{98.34} \\
        Ours & {0.071} & \underline{0.036} & \underline{0.066} & \underline{68.38} & \underline{87.06} & \underline{97.68}\\
        
        \bottomrule
    \end{tabular}
    \caption{Quantitative results on the ClearPose dataset. }
    \label{tab:clearpose_result}
    \vspace{-15pt}
\end{table}

\subsection{Ablation Study}

We evaluate three ablated variants: without the RGB encoder, without the shape prior encoder, and using only raw depth. Results in Table~\ref{tab:ablation_result} show that the shape prior is essential. The full model performs best by jointly leveraging RGB cues, metric depth, and geometric priors for improved structural consistency.

\begin{table}[htbp]
    \tabcolsep = 2.0pt
    \centering
    \renewcommand{\arraystretch}{1.2}
    \begin{tabular}{l|cccccc}
        \toprule
        \text{Method} & \text{RMSE}$\downarrow$ & \text{MAE}$\downarrow$ & \text{REL}$\downarrow$ & \text{$\delta_{1.05}$}$\uparrow$ & \text{$\delta_{1.10}$}$\uparrow$ & \text{$\delta_{1.25}$}$\uparrow$ \\
        
        \midrule
        \midrule
        w.o. RGB encoder & 0.031 & 0.010 & 0.053 & 96.77 & 98.45 & 99.19 \\
        w.o. Shape Prior Encoder & 0.033 & 0.011 & 0.054 & 93.02 & 97.44 & 99.13 \\
        w. only Raw Depth Encoder & 0.033 & 0.012 & 0.056 & 91.41 & 97.18 & 99.11 \\
        Ours & \textbf{0.030} & \textbf{0.007} & \textbf{0.047}& \textbf{97.26} & \textbf{98.59} & \textbf{99.19}\\
        \bottomrule
    \end{tabular}
    \caption{Quantitative results of the ablation study }
    \label{tab:ablation_result}
    \vspace{-15pt}
\end{table}

\subsection{Inference Speed Analysis}
We evaluate the single-frame inference latency of our method and several baselines on an NVIDIA RTX 3090 GPU with batch size 1. The image resolutions follows the original setting of these algorithms, $256\times256$ for DFNet and our approach, $224\times224$ for SwinDRNet, $504\times504$ for DA3(large), $518\times518$ for DA2(vitl) and Pi3, and $1536\times1536$ for DepthPro. As reported in Table~\ref{tab:infer_speed}, our method achieves 28.79 ms per frame, supporting real-time operation in robotic grasping scenarios, providing a favorable trade-off between inference speed and depth estimation performance.

\begin{table}[htbp]
    \tabcolsep = 2.0pt
    \centering
    \renewcommand{\arraystretch}{1.2}
    \begin{tabular}{l|ccccccc}
        \toprule
        \text{Method} & SwinDR & DFNet & DA2(vitl) & DA3(large) & Pi3 & Depth-Pro & Ours \\
        
        \midrule
        \midrule
        Infer Speed[ms] & 19.70 & 5.19 & 101.47 & 59.73 & 191.22 & 1139.89 & 28.79\\

        \bottomrule
    \end{tabular}

    \caption{Single-frame inference latency (batch size = 1) on an RTX 3090 GPU. Our method achieves real-time performance without multi-view or iterative optimization.}
    
    \label{tab:infer_speed}
    \vspace{-10pt}
\end{table}

\section{CONCLUSIONS} \label{section:conclusions}

In this work, we presented AISPO, a depth completion framework that combines multi-scale RGB-D features with an affine-invariant shape prior to improve geometric consistency for non-Lambertian objects. Trained entirely on synthetic data, AISPO generalizes effectively to real-world specular and transparent objects, producing dense, fine-grained depth maps that support reliable robotic grasping. Beyond benchmark accuracy, our experiments show that reducing catastrophic depth failures translates to substantially higher grasp success rates, particularly in scenarios involving transparent or reflective objects. By categorizing failure modes into upstream perception errors and depth estimation mismatches, we provide clearer insights into system limitations and potential improvements. Future work will focus on integrating more robust segmentation models and advanced shape-aware learning to further enhance depth reliability and manipulation performance in cluttered real-world environments.







{
\bibliographystyle{IEEEtran}
\bibliography{references}

@String(CVPR= {IEEE Conf. Comput. Vis. Pattern Recog.})

@String(ICCV= {Int. Conf. Comput. Vis.})

@String(ECCV= {Eur. Conf. Comput. Vis.})

@String(CVPR  = {CVPR})

@String(ICCV  = {ICCV})

@String(ECCV  = {ECCV})

@article{IterativeVLN,
  title={Iterative Vision-and-Language Navigation},
  author={Jacob Krantz and Shurjo Banerjee and Wang Zhu and Jason J. Corso and Peter Anderson and Stefan Lee and Jesse Thomason},
  journal={2023 IEEE/CVF Conference on Computer Vision and Pattern Recognition (CVPR)},
  year={2022},
  pages={14921-14930}
}

@misc{Grounded-SAM2,
  author = {IDEA-Research},
  title = {Grounded-SAM-2},
  year = {2024},
  howpublished = {\url{https://github.com/IDEA-Research/Grounded-SAM-2}},
  note = {Accessed: 2024-08-07}
}

@article{cleargrasp,
  title={Clear Grasp: 3D Shape Estimation of Transparent Objects for Manipulation},
  author={Shreeyak S. Sajjan and Matthew Jackson Moore and Mike Pan and Ganesh Nagaraja and Johnny Lee and Andy Zeng and Shuran Song},
  journal={2020 IEEE International Conference on Robotics and Automation (ICRA)},
  year={2019},
  pages={3634-3642}
}

@article{LIDF,
  title={RGB-D Local Implicit Function for Depth Completion of Transparent Objects},
  author={Luyang Zhu and Arsalan Mousavian and Yu Xiang and Hammad Mazhar and Jozef van Eenbergen and Shoubhik Debnath and Dieter Fox},
  journal={2021 IEEE/CVF Conference on Computer Vision and Pattern Recognition (CVPR)},
  year={2021},
  pages={4647-4656}
}

@inproceedings{transparent_net,
  title={Seeing Glass: Joint Point Cloud and Depth Completion for Transparent Objects},
  author={HaoPing Xu and Yi Ru Wang and Sagi Eppel and Al{\'a}n Aspuru-Guzik and Florian Shkurti and Animesh Garg},
  booktitle={Conference on Robot Learning},
  year={2021}
}

@article{TODE_Trans,
  title={TODE-Trans: Transparent Object Depth Estimation with Transformer},
  author={Kan Chen and Shaochen Wang and Beihao Xia and Dongxu Li and Zheng Kan and Bin Li},
  journal={2023 IEEE International Conference on Robotics and Automation (ICRA)},
  year={2022},
  pages={4880-4886}
}

@inproceedings{d3roma,
  title={D3RoMa: Disparity Diffusion-based Depth Sensing for Material-Agnostic Robotic Manipulation},
  author={Songlin Wei and Haoran Geng and Jiayi Chen and Congyue Deng and Wenbo Cui and Chengyang Zhao and Xiaomeng Fang and Leonidas J. Guibas and He Wang},
  booktitle={Conference on Robot Learning},
  year={2024}
}

@article{DDPM,
  title={Denoising Diffusion Probabilistic Models},
  author={Jonathan Ho and Ajay Jain and P. Abbeel},
  journal={ArXiv},
  year={2020},
  volume={abs/2006.11239}
}

@article{SimNet,
  title={SimNet: Enabling Robust Unknown Object Manipulation from Pure Synthetic Data via Stereo},
  author={Thomas Kollar and Michael Laskey and Kevin Stone and Brijen Thananjeyan and Mark Tjersland},
  journal={ArXiv},
  year={2021},
  volume={abs/2106.16118}
}

@article{MVTrans,
  title={MVTrans: Multi-View Perception of Transparent Objects},
  author={Yi Ru Wang and Yuchi Zhao and HaoPing Xu and Saggi Eppel and Al{\'a}n Aspuru-Guzik and Florian Shkurti and Animesh Garg},
  journal={2023 IEEE International Conference on Robotics and Automation (ICRA)},
  year={2023},
  pages={3771-3778}
}

@article{MODEST,
  title={Monocular Depth Estimation and Segmentation for Transparent Object with Iterative Semantic and Geometric Fusion},
  author={Liu, Jiangyuan and Ma, Hongxuan and Guo, Yuxin and Zhao, Yuhao and Zhang, Chi and Sui, Wei and Zou, Wei},
  journal={arXiv preprint arXiv:2502.14616},
  year={2025}
}

@article{NeRF,
  title={NeRF},
  author={Ben Mildenhall and Pratul P. Srinivasan and Matthew Tancik and Jonathan T. Barron and Ravi Ramamoorthi and Ren Ng},
  journal={Communications of the ACM},
  year={2020},
  volume={65},
  pages={99 - 106}
}

@article{DexNeRF,
  title={Dex-NeRF: Using a Neural Radiance Field to Grasp Transparent Objects},
  author={Jeffrey Ichnowski and Yahav Avigal and Justin Kerr and Ken Goldberg},
  journal={ArXiv},
  year={2021},
  volume={abs/2110.14217}
}

@inproceedings{EvoNeRF,
  title={Evo-NeRF: Evolving NeRF for Sequential Robot Grasping of Transparent Objects},
  author={Justin Kerr and Letian Fu and Huang Huang and Yahav Avigal and Matthew Tancik and Jeffrey Ichnowski and Angjoo Kanazawa and Ken Goldberg},
  booktitle={Conference on Robot Learning},
  year={2022}
}

@article{ResidualNeRF,
  title={Residual-NeRF: Learning Residual NeRFs for Transparent Object Manipulation},
  author={Bardienus Pieter Duisterhof and Yuemin Mao and Si Heng Teng and Jeffrey Ichnowski},
  journal={2024 IEEE International Conference on Robotics and Automation (ICRA)},
  year={2024},
  pages={13918-13924}
}

@article{InstantNGP,
    author = {Thomas M\"uller and Alex Evans and Christoph Schied and Alexander Keller},
    title = {Instant Neural Graphics Primitives with a Multiresolution Hash Encoding},
    journal = {ACM Trans. Graph.},
    issue_date = {July 2022},
    volume = {41},
    number = {4},
    month = jul,
    year = {2022},
    pages = {102:1--102:15},
    articleno = {102},
    numpages = {15},
    doi = {10.1145/3528223.3530127},
    publisher = {ACM},
    address = {New York, NY, USA},
}

@article{DA2,
  title={Depth Anything V2},
  author={Lihe Yang and Bingyi Kang and Zilong Huang and Zhen Zhao and Xiaogang Xu and Jiashi Feng and Hengshuang Zhao},
  journal={ArXiv},
  year={2024},
  volume={abs/2406.09414}
}

@InProceedings{Marigold,
      title={Repurposing Diffusion-Based Image Generators for Monocular Depth Estimation},
      author={Bingxin Ke and Anton Obukhov and Shengyu Huang and Nando Metzger and Rodrigo Caye Daudt and Konrad Schindler},
      booktitle = {Proceedings of the IEEE/CVF Conference on Computer Vision and Pattern Recognition (CVPR)},
      year={2024}
}

@article{Metric3D_v2,
  title={Metric3D v2: A Versatile Monocular Geometric Foundation Model for Zero-Shot Metric Depth and Surface Normal Estimation},
  author={Mu Hu and Wei Yin and China. Xiaoyan Zhang and Zhipeng Cai and Xiaoxiao Long and Hao Chen and Kaixuan Wang and Gang Yu and Chunhua Shen and Shaojie Shen},
  journal={IEEE Transactions on Pattern Analysis and Machine Intelligence},
  year={2024},
  volume={46},
  pages={10579-10596}
}

@article{RealmDreamer,
  title={RealmDreamer: Text-Driven 3D Scene Generation with Inpainting and Depth Diffusion},
  author={Jaidev Shriram and Alex Trevithick and Lingjie Liu and Ravi Ramamoorthi},
  journal={ArXiv},
  year={2024},
  volume={abs/2404.07199}
}

@inproceedings{promptDA,
  title={Prompting Depth Anything for 4K Resolution Accurate Metric Depth Estimation},
  author={Lin, Haotong and Peng, Sida and Chen, Jingxiao and Peng, Songyou and Sun, Jiaming and Liu, Minghuan and Bao, Hujun and Feng, Jiashi and Zhou, Xiaowei and Kang, Bingyi},
  journal={arXiv},
  year={2024}
}

@article{TransCG,
  title={TransCG: A Large-Scale Real-World Dataset for Transparent Object Depth Completion and A Grasping Baseline},
  author={Hongjie Fang and Haoshu Fang and Shengwei Xu and Cewu Lu},
  journal={IEEE Robotics and Automation Letters},
  year={2022},
  volume={PP},
  pages={1-8}
}

@inproceedings{clearpose,
  title={ClearPose: Large-scale Transparent Object Dataset and Benchmark},
  author={Xiaotong Chen and Huijie Zhang and Zeren Yu and Anthony Opipari and Odest Chadwicke Jenkins},
  booktitle={European Conference on Computer Vision},
  year={2022}
}

@inproceedings{dreds,
	title={Domain Randomization-Enhanced Depth Simulation and Restoration for Perceiving and Grasping Specular and Transparent Objects},
	author={Dai, Qiyu and Zhang, Jiyao and Li, Qiwei and Wu, Tianhao and Dong, Hao and Liu, Ziyuan and Tan, Ping and Wang, He},
	booktitle={European Conference on Computer Vision (ECCV)},
	year={2022}
    }

@inproceedings{zest,
  title={ZeST: Zero-Shot Material Transfer from a Single Image},
  author={Ta-Ying Cheng and Prafull Sharma and Andrew Markham and Niki Trigoni and Varun Jampani},
  booktitle={European Conference on Computer Vision},
  year={2024}
}

@inproceedings{AugmentedUnpairedData,
    title={Learning Depth Completion of Transparent Objects using Augmented Unpaired Data},
    author={Erich, Floris and Leme, Bruno and Ando, Noriaki and Hanai, Ryo and Domae, Yukiyasu},
    year=2023,
    booktitle  = {2023 IEEE International Conference on Robotics and Automation (ICRA)},
    publisher = {IEEE}
}

@article{SAID-NeRF,
  title={SAID-NeRF: Segmentation-AIDed NeRF for Depth Completion of Transparent Objects},
  author={Avinash Ummadisingu and Jongkeum Choi and Koki Yamane and Shimpei Masuda and Naoki Fukaya and Kuniyuki Takahashi},
  journal={2024 IEEE/RSJ International Conference on Intelligent Robots and Systems (IROS)},
  year={2024},
  pages={7535-7542}
}

@article{Grasp-NeRF,
  title={GraspNeRF: Multiview-based 6-DoF Grasp Detection for Transparent and Specular Objects Using Generalizable NeRF},
  author={Qiyu Dai and Yanzhao Zhu and Yiran Geng and Ciyu Ruan and Jiazhao Zhang and He Wang},
  journal={2023 IEEE International Conference on Robotics and Automation (ICRA)},
  year={2022},
  pages={1757-1763}
}

@article{SynergiesBA,
  title={Synergies Between Affordance and Geometry: 6-DoF Grasp Detection via Implicit Representations},
  author={Zhenyu Jiang and Yifeng Zhu and Maxwell Svetlik and Kuan Fang and Yuke Zhu},
  journal={ArXiv},
  year={2021},
  volume={abs/2104.01542}
}

@inproceedings{ManiSkill,
  title={ManiSkill: Generalizable Manipulation Skill Benchmark with Large-Scale Demonstrations},
  author={Tongzhou Mu and Z. Ling and Fanbo Xiang and Derek Yang and Xuanlin Li and Stone Tao and Zhiao Huang and Zhiwei Jia and Hao Su},
  booktitle={NeurIPS Datasets and Benchmarks},
  year={2021}
}

@article{DINOv2,
  title={DINOv2: Learning Robust Visual Features without Supervision},
  author={Maxime Oquab and Timoth{\'e}e Darcet and Th{\'e}o Moutakanni and Huy Q. Vo and Marc Szafraniec and Vasil Khalidov and Pierre Fernandez and Daniel Haziza and Francisco Massa and Alaaeldin El-Nouby and Mahmoud Assran and Nicolas Ballas and Wojciech Galuba and Russ Howes and Po-Yao (Bernie) Huang and Shang-Wen Li and Ishan Misra and Michael G. Rabbat and Vasu Sharma and Gabriel Synnaeve and Huijiao Xu and Herv{\'e} J{\'e}gou and Julien Mairal and Patrick Labatut and Armand Joulin and Piotr Bojanowski},
  journal={ArXiv},
  year={2023},
  volume={abs/2304.07193}
}

@article{MiDaS,
  title={MiDaS v3.1 - A Model Zoo for Robust Monocular Relative Depth Estimation},
  author={Reiner Birkl and Diana Wofk and Matthias M{\"u}ller},
  journal={ArXiv},
  year={2023},
  volume={abs/2307.14460}
}

@article{DPT,
  title={Vision Transformers for Dense Prediction},
  author={Ren{\'e} Ranftl and Alexey Bochkovskiy and Vladlen Koltun},
  journal={2021 IEEE/CVF International Conference on Computer Vision (ICCV)},
  year={2021},
  pages={12159-12168}
}

@article{swin_transformer,
  title={Swin Transformer: Hierarchical Vision Transformer using Shifted Windows},
  author={Ze Liu and Yutong Lin and Yue Cao and Han Hu and Yixuan Wei and Zheng Zhang and Stephen Lin and Baining Guo},
  journal={2021 IEEE/CVF International Conference on Computer Vision (ICCV)},
  year={2021},
  pages={9992-10002}
}

@article{ShapeNet,
  title={ShapeNet: An Information-Rich 3D Model Repository},
  author={Angel X. Chang and Thomas A. Funkhouser and Leonidas J. Guibas and Pat Hanrahan and Qi-Xing Huang and Zimo Li and Silvio Savarese and Manolis Savva and Shuran Song and Hao Su and Jianxiong Xiao and L. Yi and Fisher Yu},
  journal={ArXiv},
  year={2015},
  volume={abs/1512.03012}
}

@article{depth_anything_v2,
  title={Depth Anything V2},
  author={Yang, Lihe and Kang, Bingyi and Huang, Zilong and Zhao, Zhen and Xu, Xiaogang and Feng, Jiashi and Zhao, Hengshuang},
  journal={arXiv:2406.09414},
  year={2024}
}

@article{RefineNet,
  title={RefineNet: Multi-path Refinement Networks for High-Resolution Semantic Segmentation},
  author={Guosheng Lin and Anton Milan and Chunhua Shen and Ian D. Reid},
  journal={2017 IEEE Conference on Computer Vision and Pattern Recognition (CVPR)},
  year={2016},
  pages={5168-5177},
    url={}
}

@article{Xian2018MonocularRD_refinenet,
  title={Monocular Relative Depth Perception with Web Stereo Data Supervision},
  author={Ke Xian and Chunhua Shen and ZHIGUO CAO and Hao Lu and Yang Xiao and Ruibo Li and Zhenbo Luo},
  journal={2018 IEEE/CVF Conference on Computer Vision and Pattern Recognition},
  year={2018},
  pages={311-320}

}

@misc{moveit2025,
  author = {MoveIt Community},
  title = {MoveIt: Motion Planning Framework for Robotics},
  year = {2025},
  howpublished = {\url{https://github.com/moveit/moveit}},
  note = {Accessed: 2025-04-07}
}

@article{sobel,
  title={Computer vision and applications: a guide for students and practitioners},
  author={Bernd J{\"a}hne and Horst W. Haussecker},
  journal={Journal of Electronic Imaging},
  year={2000},
  volume={11},
  pages={115-115},
}

@article{pi3,
  title={pi3: Permutation-Equivariant Visual Geometry Learning},
  author={Wang, Yifan and Zhou, Jianjun and Zhu, Haoyi and Chang, Wenzheng and Zhou, Yang and Li, Zizun and Chen, Junyi and Pang, Jiangmiao and Shen, Chunhua and He, Tong},
  journal={arXiv preprint arXiv:2507.13347},
  year={2025}
}

@inproceedings{depth-pro,
  author     = {Aleksei Bochkovskii and Ama\"{e}l Delaunoy and Hugo Germain and Marcel Santos and
               Yichao Zhou and Stephan R. Richter and Vladlen Koltun},
  title      = {Depth Pro: Sharp Monocular Metric Depth in Less Than a Second},
  booktitle  = {International Conference on Learning Representations},
  year       = {2025},
}

@article{DA3,
  title={Depth Anything 3: Recovering the visual space from any views},
  author={Haotong Lin and Sili Chen and Jun Hao Liew and Donny Y. Chen and Zhenyu Li and Guang Shi and Jiashi Feng and Bingyi Kang},
  journal={arXiv preprint arXiv:2511.10647},
  year={2025}
}

@inproceedings{ClearDepth,
  title={ClearDepth: Efficient Stereo Perception of Transparent Objects for Robotic Manipulation},
  author={Kaixin Bai and Huajian Zeng and Lei Zhang and Yiwen Liu and Hongli Xu and Zhaopeng Chen and Jianwei Zhang and Camera Parameters},
}

@article{StereoAnything,
  title={StereoAnything: Advanced Zero-Shot Stereo Imaging for Robotic Grasp Detection With Transparent Objects.},
  author={Kaixin Bai and Lei Zhang and Yiwen Liu and Zhaopeng Chen and Jianwei Zhang},
  journal={IEEE transactions on cybernetics},
  year={2026},
  volume={PP},

}
}

\end{document}